\documentclass[letterpaper, 10 pt, conference]{ieeeconf}  
\usepackage{times}

\usepackage{amsthm, amsfonts, amsmath, amssymb, mathtools}
\usepackage{multicol}
\usepackage{xcolor}
\usepackage[bookmarks=true, hidelinks]{hyperref}
\usepackage{subcaption}
\usepackage{tcolorbox}
\tcbuselibrary{listings, breakable}
\usepackage{booktabs}
\usepackage{listings}
\usepackage{cuted}    
\usepackage{caption}

\IEEEoverridecommandlockouts                              

\title{\LARGE \bf
Using Language Models as Closed-Loop High-Level Planners for Robotics Applications: A Brief Overview and Benchmarks}

\author{Hao Wang$^{1, 2}$, Sathwik Karnik$^{2}$, Beatrice Lim$^{3}$, and Somil Bansal$^{2}$ 
\thanks{This research is supported in part by the DARPA Assured Neuro Symbolic Learning and Reasoning (ANSR) program and by the NSF CAREER program (2240163).}%
\thanks{$^{1}$Hao Wang is with the Ming Hsieh Department of Electrical and Computer Engineering, University of Southern California.
        {\tt\small haowwang@usc.edu}}%
\thanks{$^{2}$The authors are with the Department of Aeronautics and Astronautics, Stanford University.
        {\tt\small \{haowwang, sathwik, somil\}@stanford.edu}}%
\thanks{$^{3}$Beatrice Lim is with the Department of Mechanical Engineering, Stanford University.
        {\tt\small bealim@stanford.edu}}%
}
\newtheorem{note}{Note}
\newtheorem{problem}{Problem}
\newtheorem{definition}{Definition}

\newcommand{\cubeeasy}{\textsc{Cube-Easy}}
\newcommand{\ycbeasy}{\textsc{YCB-Easy}}
\newcommand{\ycbmedium}{\textsc{YCB-Medium}}
\newcommand{\ycbhard}{\textsc{YCB-Hard}}
\newcommand{\clfull}{\textsc{CL-Full}}
\newcommand{\clshort}{\textsc{CL-Short}}
\newcommand{\clhalf}{\textsc{CL-Half}}

\begin{document}

\maketitle

\thispagestyle{empty}
\pagestyle{empty}

\renewcommand{\thefootnote}{*}
\begin{abstract}%
Large Language Models (LLMs) and Vision Language Models (VLMs) have become popular tools for embodied high-level planning. However, their deployment in black-box settings often leads to unpredictable or costly errors. To harness their capabilities more reliably in robotic systems, we empirically investigate practical strategies for integrating language models as closed-loop planners. Concretely, we study how the \emph{control horizon} and \emph{warm-starting} impact the performance of language model-based planners. We design and conduct controlled experiments to extract actionable insights, providing recommendations that can help improve the performance and robustness of language model-based embodied planning. The full implementation and experiments are available on the project website\footnote{\url{https://github.com/haowwang/vlm_embodied_planning_benchmarking}}.
\end{abstract}

\section{Introduction}
Large Language Models (LLMs) and Vision Language Models (VLMs) have shown impressive open-world reasoning capabilities \cite{gpt4_report, gemini_2_5, deepseek_r1}, and they have proved to be powerful tools for high-level planning for robotic applications \cite{huang2022_llm_planner, driess2023_palm_e, shi2025_hi_robot}. Specifically, such planning tasks involve translating a high-level task instruction (e.g., ``make a salad") into a sequence of abstract sub-tasks (e.g., ``take the lettuce out from the fridge" and ``put the cutting board on the counter"). High-level planning differs from low-level planning; rather than directly producing actionable robot actions like end-effector poses or joint torques, a high-level planner generates abstract sub-tasks that are required to be processed by low-level planners into physical actions. High-level planning is a salient research direction as complex real world tasks typically require decomposition into more manageable sub-tasks that can be effectively translated into executable robot actions. 

Despite recent progress in using these models for high-level planning in robotic applications, how to utilize these models as \emph{closed-loop} high-level planners remains an active area of research. In this work, we aim to understand how to effectively leverage LLMs/VLMs as closed-loop high-level planners from an empirical and practical perspective in a black-box setting. We pose the problem of closed-loop high-level planning in the framework of Model Predictive Control (MPC) \cite{garcia1989_mpc}, as its repeated process of planning and execution until task completion has striking similarities with how LLMs/VLMs are used for high-level planning. Under this framework, the planning problem is conceptualized as an optimal control problem, where the instructions provided in the prompt serve as the objective, constraints implicitly or explicitly specified in the prompts are the constraints, and the language model's internal knowledge of the physical world acts as the dynamics constraint. This optimal control problem is repeatedly solved by the language model, one or more proposed actions are executed by the robot, and this process of planning-execution continues until the requested task is completed. 

In this work, we first provide a brief overview of how LLMs/VLMs are used as high-level planners for robotics applications in the existing literature. Then, we conduct controlled benchmarking experiments to investigate the impact of two fundamental characteristics of closed-loop planners: frequency of replanning (or equivalently, control horizon) and warm-starting. We provide observations and practical recommendations, based on empirical results, on how to effectively utilize LLMs/VLMs as high-level planners for robotic applications.  

\section{Overview of Language Model-Based High-Level Planners for Robotics Applications}

\subsection{Language Models for High-Level Embodied Planning}

Roughly speaking, embodied planning is planning for robots with physical embodiment situated in the physical world, taking into consideration the robot's embodiment and its surrounding environment. The seminal work of Huang et al. \cite{huang2022_llm_planner} demonstrates that LLMs could generate correct, executable plans in a zero-shot manner by simply prompting the language model to break down tasks step-by-step. However, to ensure the generated plans are physically executable by a specific robot, the language model needs to be \emph{grounded}, meaning that the language model needs to consider the robot's physical embodiment and/or surrounding environment during planning.

SayCan \cite{ahn2022_saycan} achieves this by having the language model score the feasibility of various textual sub-tasks, which are then multiplied by the physical affordance score of the corresponding skill at the current state, effectively using the LLM as a semantic value function for high-level action selection. Alternatively, rather than outputting flat sequences of text, another line of research utilizes language models to output executable code, providing planning loops, conditionals, and parameterization that a flat action sequence cannot express. Some early works in this direction include Code as Policies \cite{code_as_policies}, PROGPROMPT \cite{prog_prompt}, and Instruct2Act \cite{instruct2act}. A third approach uses the language models to bridge the gap between natural language and formal, verifiable planning constraints, instead of generating a textual plan. In works like LLM+P \cite{llmp}, AutoTAMP \cite{autotamp} and VLM-TAMP \cite{yang2025_vlm_tamp}, the language model translates open-vocabulary user instructions and environment descriptions into formal representations, such as PDDL \cite{mcdermott1998pddl} or Signal Temporal Logic \cite{signal_temporal_logic}. These formal specifications are then passed to classical planners or verifiable executors. In this paradigm, the LLM acts as a semantic translator, while the actual task of combinatorial search and ensuring physical/logical feasibility is offloaded to classical task planners. More recently, the incorporation of vision as an additional modality has further extended grounding capabilities, enabling planners to reason directly from visual observations of the environment \cite{driess2023_palm_e, hu2023_vlm_planning}, marking the transition from purely language-based to multimodal high-level planners.

\subsection{Embodied Chain of Thought and Reasoning}

The integration of language models into robotics initially treated the models as direct observation-to-action mapping. However, the community quickly discovered that forcing a model to directly output high-level plans without intermediate deliberation often leads to brittle performance. To address this, recent works seek to incorporate the concept of Chain-of-Thought (CoT) prompting \cite{wei_cot}, which originally demonstrated that eliciting step-by-step reasoning significantly improves a language model's ability to solve complex problems. 

Formalized by frameworks like ReAct \cite{yao2022react}, the approach of  interleaving of reasoning and acting demonstrates that allowing a model to generate intermediate reasoning steps before committing to an action significantly improves its performance in practice. This reasoning loop is further enriched by systems like Inner Monologue \cite{huang2022_inner_monologue}, which blends CoT reasoning with environmental feedback by injecting execution successes, failures, and scene changes directly into the model's ongoing internal reasoning chain. 

As embodied planners transition to multimodal inputs, this reasoning process must explicitly ground itself in visual and spatial realities, as standard semantic CoT often fails to respect physical constraints. To bridge this gap, recent works have formalized \emph{Embodied Chain-of-Thought} (ECoT) \cite{ecot}, which explicitly requires the model to output a structured sequence of physical reasoning steps before committing to an action. This ensures that the intermediate reasoning trace is not just semantically sound, but physically executable, substantially improving the model's performance.

\subsection{Closed-Loop High-Level Embodied Planning with Language Models}

While the paradigms discussed in the previous section establish how language models can generate grounded plans, relying on open-loop execution in the real world is difficult due to the following reasons: 1) language models do not always generate correct high-level plans, 2) even under the correct high-level plans, low-level execution is prone to failure, and 3) the surrounding environment can change unpredictably. As a result, the planner must be able to re-generate high-level plans online as the robot executes actions and interacts with the environment. 

Early approaches to closing this planning-execution relied primarily on text or code. For instance, Inner Monologue \cite{huang2022_inner_monologue} demonstrates that feeding text-based execution successes, failures, and updated scene descriptions back into the prompt allows the language model to replan and recover from errors. Subsequent works have expanded on text-driven replanning by incorporating few-shot grounded planning \cite{song2023_llm_planner} and explicit motion failure reasoning \cite{wang2024_llm3} to better help the model diagnose execution errors. Similarly, when using code as an intermediate representation, runtime tracebacks and execution errors can be parsed and returned to the language model to iteratively debug and refine the planning logic, as shown in PROGPROMPT \cite{prog_prompt}. Beyond external feedback, recent literature also explores self-correction, where models assess task progress through self-questioning \cite{shin2025_socratic_planner} or multi-stage reflection \cite{feng2025_reflective_planning}.

Building upon the multi-modal capabilities discussed previously, recent VLM-based planners can natively close the planning-execution loop by directly ingesting sequential visual observations. This allows the model to visually assess task progress and replan without relying on external text-translation systems \cite{driess2023_palm_e}. Specifically, planners such as Replan \cite{skreta2024_replan}, ReplanVLM \cite{mei2024_replanvlm}, and hierarchical vision-language-action models like Hi Robot \cite{shi2025_hi_robot} demonstrate how direct visual perception can be utilized to dynamically adjust plans in a closed-loop manner. 

While the methods mentioned above are diverse, they share one commonality: the LLMs/VLMs are used in a \emph{closed-loop} fashion. The vast majority of the methods perform replanning in one of two ways: 
\begin{enumerate}
    \item replan after 1 planned action is executed by the system \cite{ahn2022_saycan, hu2023_vlm_planning, feng2025_reflective_planning,zhang2023_grounding_tamp}
    \item replan only if a planned action fails to execute or requires intervention  \cite{driess2023_palm_e, shi2025_hi_robot, mei2024_replanvlm, wang2024_llm3, skreta2024_replan, shin2025_socratic_planner, yang2025_vlm_tamp, mei2024_replanvlm}. 
\end{enumerate}
The replanning frequency (or equivalently, control horizon) of the first approach is 1, and the second approach has a replanning frequency equals to the length of the task. Selection of the control horizon is often an afterthought in the literature, as existing works do not justify the choice of control horizon or explore its impacts on the performance of the high-level planner. It is generally believed that more frequent replanning (shorter control horizon) leads to more robust behaviors and better performance; however, it is unclear whether this intuition holds for language model-based high-level planners, as they have limited reasoning capabilities and can be prone to errors.

\section{Problem Formulation}
In this work, we investigate the use of language models as high-level planners for robotic applications empirically. We will first define relevant terminologies: open-loop planner, closed-loop planner, control horizon, and warm-starting in the context of high-level planning. 

\begin{definition}[Closed-loop Planner]
A closed-loop planner generates a plan at its instantiation and generates at least one additional plan (i.e., replanning) in subsequent time steps based on updated information about the system and the environment, whereas an open-loop planner does not generate any additional plans beyond the initial plan.
\end{definition}

\begin{definition}[Control Horizon]
The control horizon of a closed-loop planner is the number of actions designated to be executed by the system before a new plan is regenerated based on updated information about the system and the environment. The shorter the control horizon, the more frequently the planner replans. 
\end{definition}

\begin{definition}[Warm-Starting]\label{def:ws}
Generally, warm-starting is the practice of providing a planner with initialization. In this work, warm-starting specifically refers to providing a closed-loop planner with a previously generated plan, and potentially along with other feedback information, as initialization when it replans. 
\end{definition}

More specifically, we are interested in the following problems:

\begin{problem}\label{prob:cl}
    What are the benefits of using language models in a closed-loop manner as opposed to in an open-loop manner?
\end{problem}

\begin{problem}\label{prob:ch}
    How does the control horizon impact the performance of closed-loop language model-based high-level planners?
\end{problem}

\begin{problem}\label{prob:ws}
    How does warm-starting impact the performance of closed-loop language model-based high-level planners?
\end{problem}

\begin{note}
    \textcolor{black}{While the planning horizon (i.e., the number of actions contained within each generated plan) is an essential parameter in closed-loop planning, LLM/VLM planners typically produce complete task-length plans at every iteration; thus, we do not investigate the impacts of the planning horizon in this work.}
\end{note}

We situate our investigation in the context of using a hierarchical planning and control framework (high-level planning + low-level motion planning) to solve long-horizon manipulation tasks, for the following reasons: 1) this task setup typically requires multi-step reasoning, including geometric and logical reasoning, and 2) long-horizon manipulation is a salient open research problem and can potentially unlock numerous downstream applications. In this setup, the language model acts as the high-level planner, while the motion planner serves as the low-level controller. The language model is provided with a set of available primitive actions (e.g., pick and place), and it is expected to generate a plan (a sequence of primitive actions) that accomplishes the defined task. The motion planner then translates each primitive action in the plan into a trajectory in the robot's configuration space. To ensure safety and feasibility, the motion planner performs continuous collision checks between the robot arm and all objects in the scene, automatically rejecting trajectories that would result in unintended contact during the ``pick'' or ``place'' phases. It is worth noting that while the choice to use a motion planner as the low-level controller limits the options of primitive action, motion planners are generally robust and allow us to focus this work on high-level planning instead of low-level control.

\section{Benchmarking Experiments}
\begin{figure*}
    \centering
    \includegraphics[width=0.8\textwidth]{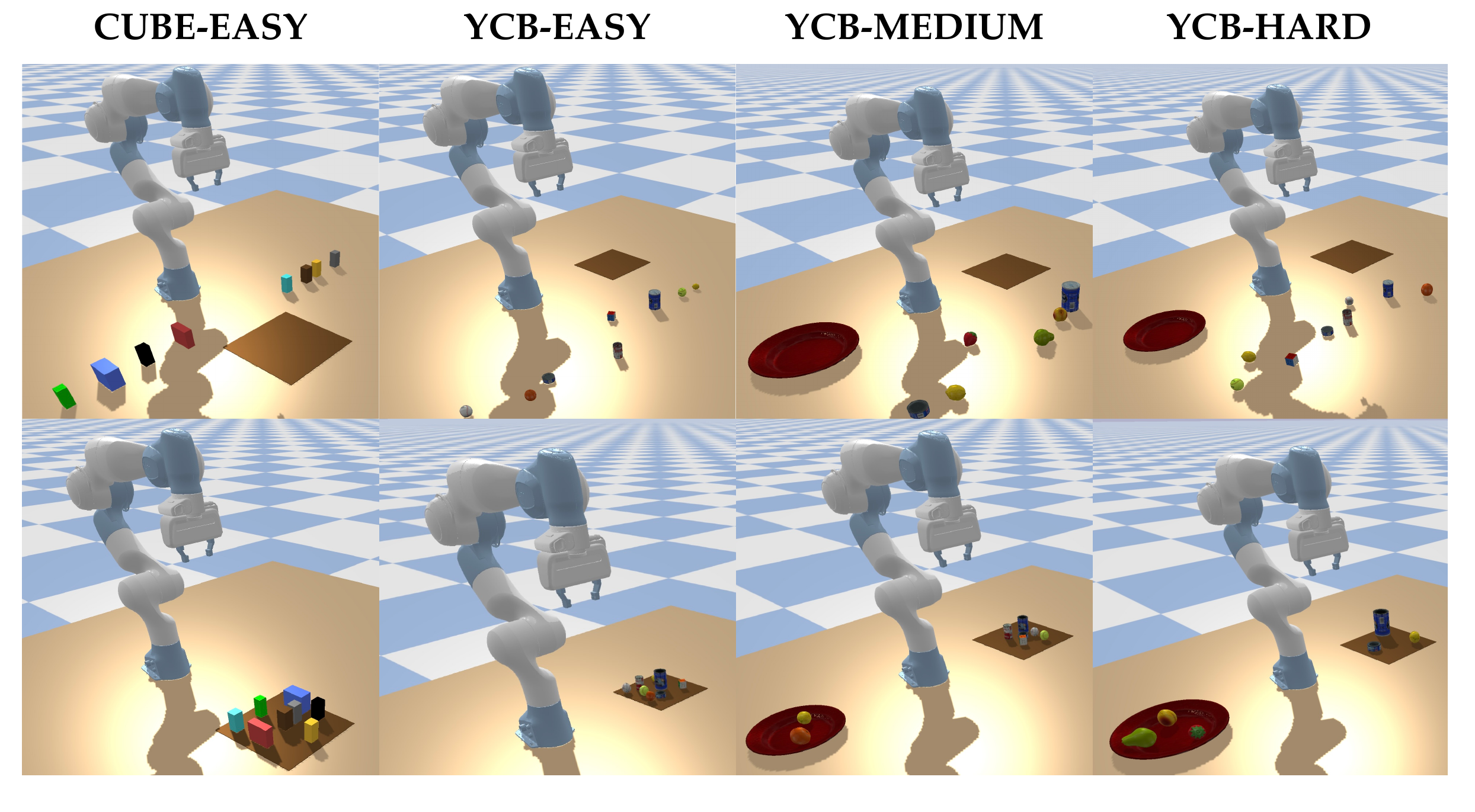}
    \captionof{figure}{This figure captures the initial (top) and final observation images (bottom) for sampled trajectories in the 4 environments: $\cubeeasy$, $\ycbeasy$, $\ycbmedium$, and $\ycbhard$.}
    \label{fig:task_envs}
\end{figure*}

In this work, we consider 4 task environments with increasing demand on the reasoning capabilities of the language models, more specifically VLMs. All task environments involve rearranging objects in the scene using 2 primitive actions: \texttt{pick(object)} and \texttt{place(object, location)}, with \emph{logical} constraints imposed on the order in which certain objects are moved. We utilize 2 planners: 1) \textbf{open-loop planner (OL)}, and 2) \textbf{closed-loop planner (CL)}, which replans after $N$ action(s) of the previously generated plan are executed (i.e., the control horizon is $N$) or an action from the previously generated plan fails to execute. Practically speaking, each planner is simply a way of using a VLM to generate a sequence of primitive actions. The prompts used for each planner are slightly different from each other, but they all contain: 1) a description of the task, including \emph{requirements} and \emph{constraints}, 2) information about the robot embodiment, including formats and required parameters of available primitive actions, and 3) current location of objects in the scene. By default, CL planners utilize warm-starting. The planner prompts are provided in Appendix \ref{appen:planner_prompts}.  

The spatial information of objects is included to ensure that the VLM can reason about geometric feasibility and provide the metric parameters (i.e., specific coordinates) necessary for the motion planner to execute the ``place'' primitive, thus grounding the VLM's output in the workspace. Moreover, we found empirically that the benchmarked VLMs do not demonstrate the necessary capabilities for object position estimation and reasoning with images alone.

In our experiments, all planners are evaluated with 3 different VLMs: GPT-4.1-mini, Gemini-2.5-flash, and
Llama-4-Maverick-17B. Across all VLMs used, the temperature hyperparameter is set to 0.0, denoting a greedy decoding scheme. All VLMs used assume that images are provided as PNGs with no applied resolution adjustments or resizing. The VLM's textual responses are processed by a parsing script that utilizes regular expressions to extract the intended sequence of primitive actions and their metric parameters. If a response cannot be successfully parsed into the required action format, it is treated as a generation failure.

For all task environments, we randomly generate 50 initial conditions (i.e., 50 randomized trials) and have planners perform high-level planning from each of these initial conditions using each of the 3 VLMs. To ensure fair comparisons between planners, for each initial condition, the CL planners utilize the output of the OL planner as the initial plan. 

\subsection{Implementation Details for VLM Planners}
\textcolor{black}{Recall that for Problem \ref{prob:cl}, our goal is to investigate the benefit of using CL planners over OL planners. For this problem, we compare the OL planner against a CL planner with control horizon equal to the length of the task defined by the task environment. This CL planner effectively only replans if a primitive action fails to execute.} 

In Problem \ref{prob:ch}, we aim to determine how the control horizon impacts the performance of CL planners. Let $k$ denote the length of the task for a specific task environment, or equivalent, $k$ is the minimum number of primitive actions required to complete the task. We consider CL planners with 3 different control horizons: 1) short, the planner replans every 2 steps, 2) half, the planner replans every $\frac{k}{2}$ steps, and 3) full, the planner replans every $k$ steps. 

Lastly, in Problem \ref{prob:ws}, we assess the effects of warm-starting on the performance of CL planners. We utilize CL planners with 2 different control horizons: 1) short and 2) half, and their non-warm-starting counterparts. The non-warm-starting variant is identical to its counterpart, but omits the warm-starting information from the prompt.

For any CL planners, we limit the total number of VLM queries (i.e., the number of opportunities to produce a plan) to $\frac{2k}{\text{control horizon}}$ for fair comparisons. We do so with the consideration that VLMs are prone to mistakes, and more VLM queries could lead to more mistakes being committed. To avoid artificially disadvantaging CL planners with shorter control horizons, we provide them with more available VLM queries (i.e., opportunities to replan). 

\subsection{Task Environments}

We evaluate the planners in 4 environments: $\cubeeasy$, $\ycbeasy$, $\ycbmedium$, and $\ycbhard$, covering various degrees of complexity in reasoning and planning. While $\cubeeasy$ focuses on manipulating cube objects, \textcolor{black}{the YCB environments, named after the YCB Object and Model Set \cite{ycb}, contain everyday household and kitchen objects for additional variety.} The task environments are visualized in Fig. \ref{fig:task_envs}, and we describe the environments as follows, with additional details of the prompts for each environment in Appendix~\ref{appen:env_prompts}.

\begin{itemize}
    \item $\cubeeasy$ tasks the robot with placing differently colored boxes into the basket (the square-shaped brown region). This task evaluates the basic reasoning and planning capabilities of the planner.
    \item $\ycbeasy$ tasks the robot with placing different YCB objects into the basket. This task only differs from $\cubeeasy$ by replacing the boxes with YCB objects.
    \item $\ycbmedium$ tasks the robot with placing fruit on the red plate and other objects into the basket. $\ycbmedium$ provides another level of complexity over $\ycbeasy$ by requiring the planner to reason about placing objects in multiple goal locations.
    \item $\ycbhard$ tasks the robot with making a fruit salad, requiring the robot to place certain fruit on the plate and the unused objects in the basket. $\ycbhard$ differs from $\ycbmedium$ by providing additional descriptors in the prompts, such as ``sour fruit.'' Thus, $\ycbhard$ challenges the planner to reason about the properties of objects. $\ycbhard$ presents the most difficult tasks to evaluate both reasoning and planning performance.   
\end{itemize}

It is crucial to note that all 4 task environments are \emph{static}: objects remain stationary unless directly manipulated by the robot, and no new objects are introduced during task execution. We do not analyze dynamic environments, as they naturally favor CL planners with shorter control horizons over those with longer horizons, and this would obscure our investigation of how the control horizon impacts both the \emph{geometric} and \emph{logical} reasoning capabilities of CL planners. 

Furthermore, we note that the explicit logical constraints provided in our task prompts (e.g., spatial relations or containment) often induce \emph{implicit constraints}. For instance, a requirement to place an object in a specific region implicitly constrains the reachable configuration space for subsequent actions and may impose a strict ordering on the plan to ensure that earlier placements do not create geometric obstructions. This coupling of explicit and implicit constraints increases the reasoning complexity for the VLM planner.

\subsection{Metrics}

For any particular metric, the comparison is conducted between planners employing the same VLM within the same task environment, and we refer to each such setting of VLM and task environment as a \emph{scenario}. At the most fundamental level, we try to draw conclusions for Problems \ref{prob:cl} - \ref{prob:ws} from our experiments by answering questions of the following format: 
\begin{center}
\emph{``Is Planner A better than Planner B for metric X"}.
\end{center}
for relevant planners and metrics using all available scenarios. In this work, answers to such questions are binary, and metrics of interest are population statistics. Hence, we utilize hypothesis testing, more specifically, a two-proportion z-test, to determine whether the answer to such a question is statistically significant with significance level of $0.05$ throughout this work. 

The primary metric we utilize to evaluate the overall performance of a planner is the \textbf{Task Completion Rate}, which notes the percentage of trials in a scenario where all objects are placed successfully and all logical constraints are satisfied. In addition to this primary metric, we also devise secondary metrics to assess the \emph{geometric} and \emph{logical} reasoning performance of a planner. 

To evaluate geometric reasoning, we use \textbf{Goal Achieved Rate}, the percentage of trials in a scenario where all objects are placed successfully without considering the logical constraints. The Goal Achieved Rate reflects the geometric reasoning performance of the planner, as the planner is tasked with placing objects in designated locations without colliding with other objects by selecting the \texttt{location} parameter, $(x,y,z)$ coordinate of the proposed placement, of the \texttt{place} primitive action. By construction, the Task Completion Rate is upper-bounded by the Goal Achieved Rate.  

We use 3 metrics to assess the logical reasoning performance of a planner. The first metric is the \textbf{Correct Final Logical Plan Rate}, which is the percentage of trials in a scenario that satisfy all logical constraints. This metric evaluates the overall logical reasoning performance of a planner over a scenario. The following two metrics, \textbf{Positive/Negative Logical Correction Rates}, assess the logical reasoning performance at a more granular level. A positive logical correction is made when the planner decreases the number of logical violations from the previous planning iteration, whereas a negative logical correction is made when the number of logical violations increases. The Positive/Negative Logical Correction Rates are fractions of positive/negative logical corrections suggested by the planner over the number of all possible positive/negative corrections. 

We provide all the results for each metric for all task environments, VLMs, and planners in Appendix \ref{appen:detailed_results}.

\section{Results}
In this section, we discuss experiment results and observations for Problems \ref{prob:cl}-\ref{prob:ws} in subsections \ref{subsec: obs_cl} - \ref{subsec:ws}, respectively, and we briefly discuss our overall recommendations for utilizing VLMs as closed-loop high-level planners in subsection \ref{subsec:rec}. 

\begin{figure}[t!]
    \centering
    \includegraphics[width=1\linewidth]{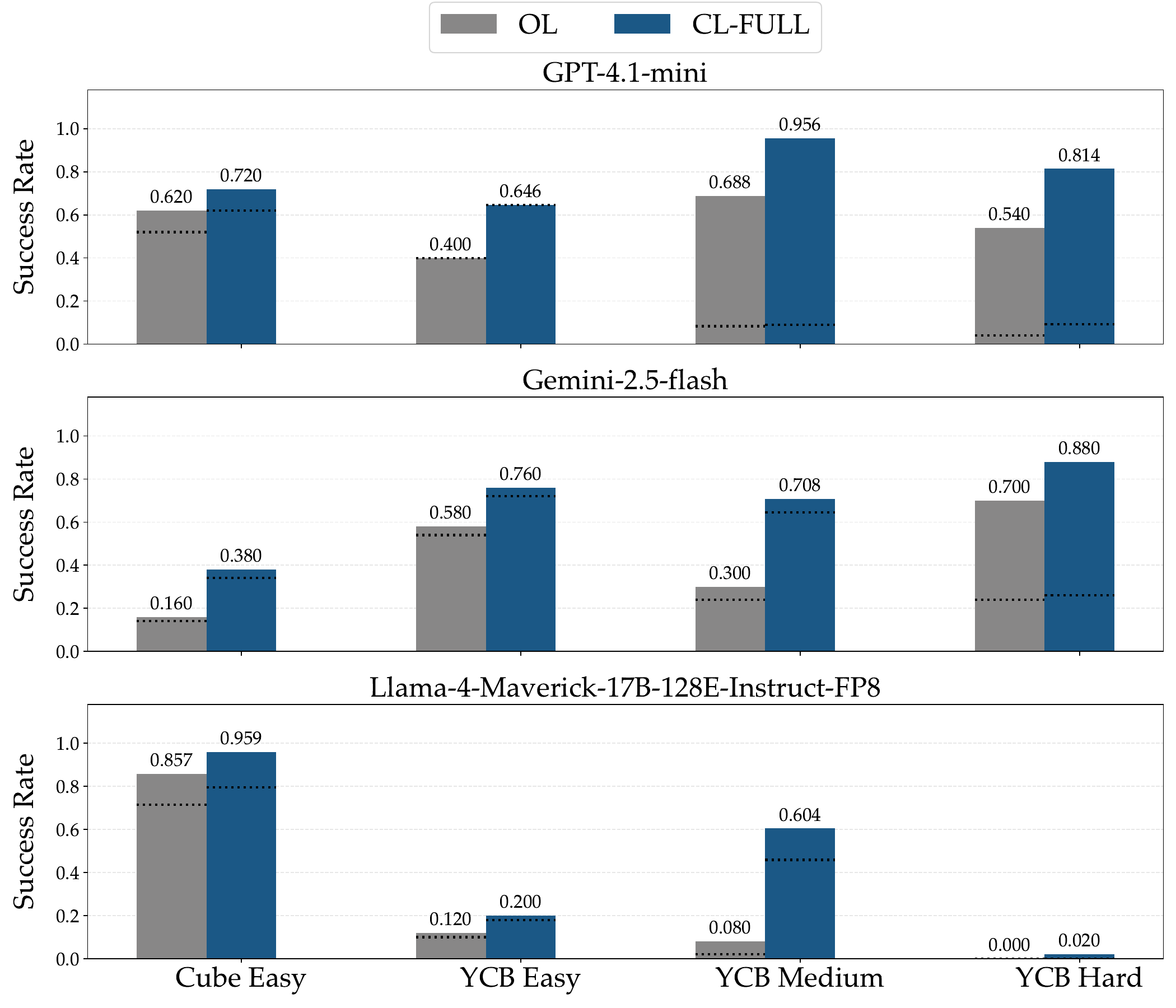}
    \caption{Goal Achieved Rates and Task Completion Rates for the OL and $\clfull$ planner. Each bar indicates the Goal Achieved Rate, and the black dotted line indicates the Task Completion Rate.}
    \label{fig:hypo_1}
\end{figure}

\subsection{Observation 1: Closed-Loop High-Level Planning is Preferred Over Open-Loop}\label{subsec: obs_cl}
\textcolor{black}{For most robotic applications, if computation permits, closed-loop is preferred over open-loop for various reasons. One of the most crucial reasons is that the environment changes as the robot interacts with it, necessitating replanning using updated information. In our experiments, we found that closed-loop is still preferred even when the environment is \emph{static}, largely because VLMs do not always generate goal-achieving and constraint-satisfying plans.}

We first discuss the geometric reasoning performance of the planners. In Fig. \ref{fig:hypo_1}, we present the Goal Achieved Rate and Task Completion Rate of the open-loop (OL) planner and a closed-loop (CL) planner. In 4 environments and 3 VLMs (a total of 12 scenarios), the OL planner does not attain 100\% Goal Achieved Rate in any scenario, and replanning consistently helps in achieving the goal more often. The CL planner achieves higher Goal Achieved Rates in all 12 scenarios, with an average improvement of $21.7\%$, and 7 of the 12 results are statistically significant. Unsurprisingly, the CL planner outperforms the OL planner in Goal Achieved Rate, as the CL planner is given an extra opportunity to correct the misplacement that causes the OL planner to fail. 

On the other hand, the CL planner does not clearly demonstrate superior logical reasoning performance over the OL planner. The CL planner achieves a higher Correct Final Logical Plan Rate in 1 of the 12 scenarios, and the OL planner performs better in 3 of the 12 scenarios. However, none of the results are statistically significant. It is important to note that this CL planner only has one replanning opportunity, and as a result, it is unable to correct most logical reasoning errors. This is largely why, in several scenarios shown in Fig. \ref{fig:hypo_1}, such as GPT in $\ycbmedium$ and $\ycbhard$, and Gemini in $\ycbhard$, the CL planner substantially outperforms the OL planner in Goal Achieved Rate, but they have similar Task Completion Rates. 

For robotic applications, the conventional wisdom is that closed-loop planners are more robust to changes in the environment. With Observation~\ref{subsec: obs_cl}, we argue that closed-loop \emph{VLM} planners are also more robust to errors during VLM inference, as closed-loop planners have more opportunities to replan and correct their errors. 

\begin{figure}[t!]
    \centering
    \includegraphics[width=1\linewidth]{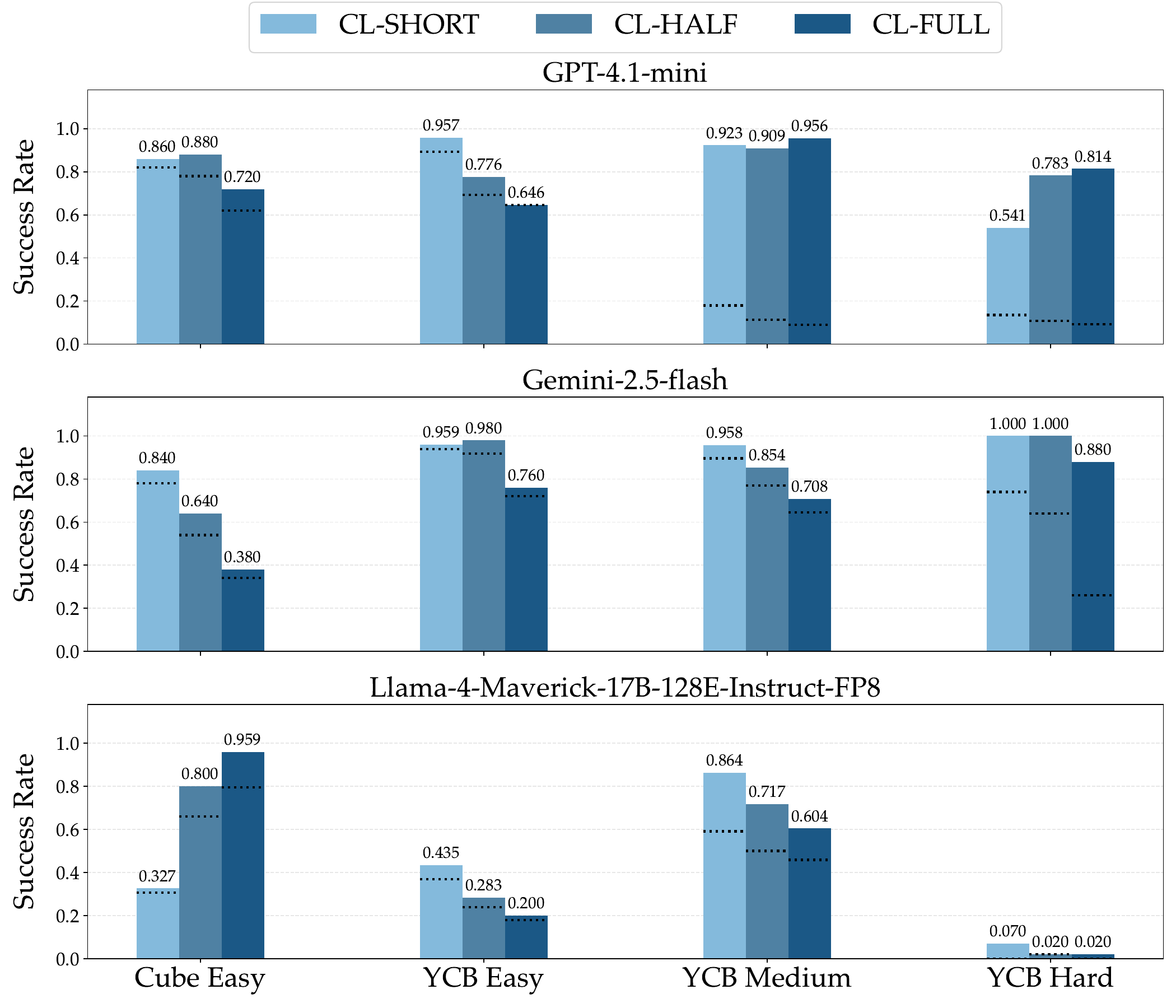}
    \caption{Goal Achieved Rates and Task Completion Rates for 3 CL planners with different control horizon settings.}
    \label{fig:hypo_2}
    \vspace{-1em}
\end{figure}

\subsection{Observation 2: Shorter Control Horizon is Not Always Better}
Intuitively, more frequent replanning (a shorter control horizon) should be beneficial as the planner has more opportunities to reason about the task with updated information about the environment, leading to better performance in both geometric and logical reasoning. However, in our experiments, we found that the CL planner with the shortest control horizon is not clearly better than CL planners with longer control horizons. Recall that we compare CL planners in 3 control horizon settings: short ($\clshort$), half ($\clhalf$), and full ($\clfull$).

In Fig. \ref{fig:hypo_2}, we plot the Goal Achieved Rate and Task Completion Rate for 3 CL planners with 3 VLMs in 4 environments. Out of 12 scenarios, $\clshort$ achieves the best Task Completion Rate in 10 of the scenarios, but only 2 of the results are statistically significant, suggesting that a shorter control horizon is not meaningfully beneficial for the overall task performance. For geometric reasoning performance, $\clshort$ achieves the best Goal Achieved Rate in 6 of the scenarios, but again, only 2 of the results are statistically significant, indicating that having a shorter control horizon does not meaningfully improve the geometric reasoning performance of CL planners. 

As for the logical reasoning performance, $\clshort$ achieves the best Correct Final Logical Plan Rate in 9 out of 12 scenarios, but only one of the results is statistically significant. Similar trends hold true for the Positive and Negative Logical Correction Rates. Only 2 results are statistically significant for the Positive Logical Correction Rate, and none of the results are statistically significant for the Negative Logical Correction Rate, indicating that there is no particular control horizon that works best in terms of logical reasoning performance. While a shorter control horizon affords the CL planner more replanning opportunities and more opportunities for positive logical correction, it also provides more opportunities for the VLM to commit negative logical corrections. As a result, we believe that the logical reasoning capability of a CL planner is meaningfully impacted by the VLM used in the planner, rather than the control horizon.

\subsection{Observation 3: Warm-Starting is Typically Beneficial}\label{subsec:ws}
In this work, to generate plans with warm-starting (see Def.~\ref{def:ws}), we provide both the previously generated plan and the execution statuses of primitive actions of the previously generated plan to the CL planners, similar to the practice in \cite{wang2024_llm3}. In our experiments, we found that warm-starting is consistently beneficial to the overall performance, as well as the geometric and logical reasoning performance, of the CL planners. 

\begin{figure}[t!]
    \centering
    \includegraphics[width=1\linewidth]{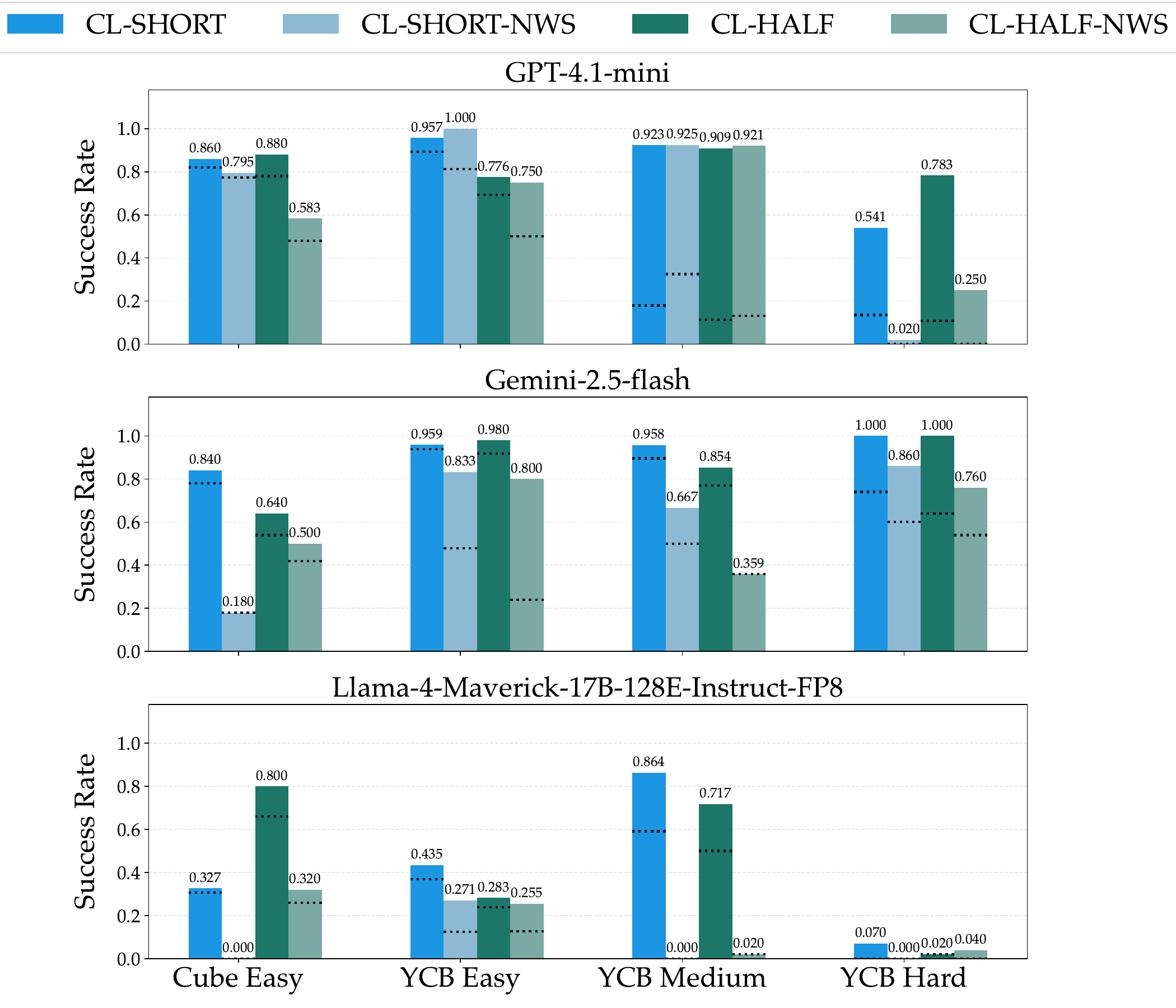}
    \caption{Goal Achieved Rate and Task Completion Rate of warm-starting and non-warm-starting variants of CL planners with 2 control horizon settings.}
    \label{fig:hypo_3_1}
\end{figure}

We first discuss the overall performance of the CL planners. In Fig. \ref{fig:hypo_3_1}, we plot the Task Completion Rates for the warm-starting and non-warm-starting variants of the $\clshort$ and $\clhalf$ planners. Out of the 24 scenarios (4 environments, 3 VLMs, and 2 CL planners with different control horizon settings), the warm-starting variants are better in 21 out of 24 scenarios (13 of the 21 results are statistically significant) with an average improvement of $28.2\%$. On the other hand, the non-warm-starting variants are better in 3 scenarios, and none of the results are statistically significant. The results indicate that warm-starting is beneficial to the overall performance of CL planners. 

Unsurprisingly, warm-starting is also beneficial to the geometric reasoning capabilities of CL planners, as the warm-starting variants have higher Goal Achieved Rates, which are shown in Fig. \ref{fig:hypo_3_1}, than their non-warm-starting counterparts, with an average improvement of $31.7\%$, in 20 scenarios (14 of the 20 results are statistically significant). It is worthwhile to note that we provide the execution statuses of actions from the previously generated plan. As a result, when CL planners replan due to execution failures, they have explicit feedback about the failures. We believe this information is important to improving the Goal Achieved Rate, as it informs the planners where and how the previously generated plan fails. 

It is worthwhile to point out that in some scenarios, for instance, GPT in $\ycbhard$, Gemini in $\cubeeasy$, and Llama in $\ycbmedium$, the performance of non-warm-starting variants of the CL planners completely collapses, clearly suggesting that the CL planners in those scenarios would not work without warm-starting. This result is expected, given that without warm-starting, each replanning attempt is an independent query to the VLM, and the probability of stringing together successful primitive actions decreases geometrically, and this probability decreases as the task becomes longer and the planner utilizes more frequent replanning. 

For logical reasoning performance, we found that warm-starting helps CL planners make fewer negative logical corrections. The warm-starting CL planners have a lower Negative Logical Correction Rate in 21 out of the 24 scenarios, with an average decrease of $7.0\%$, and 14  of the 21 results are statistically significant. The non-warm-starting CL planners performed better in 3 of the 24 scenarios, but none of the 3 results is statistically significant. On the other hand, we found that the warm-starting variants of CL planners are less likely to make positive logical corrections than their non-warm-starting counterparts. In 16 out of 24 scenarios, with 6 of the results statistically significant, the non-warm-starting variants have higher Positive Logical Correction Rates. By warm-starting, the planner is biased towards the previously generated plan. Though we explicitly instruct the planner to reason about both the task and the previously generated plan in the prompt, it is unsurprising that the planner makes fewer positive logical corrections, as it is more likely to adhere to the previously generated incorrect plan than in the situation where warm-starting is not utilized. 

\subsection{Recommendations for Using VLMs as Embodied High-Level Planners}\label{subsec:rec}
We summarize our most important recommendations: utilize closed-loop planners with warm-starting whenever possible, and choose a control horizon so that the system is sufficiently reactive for the task (but no need to replan as frequently as computation permits). Certain VLMs are more effective at high-level planning for robotic applications than others, and the overall performance of the planner is heavily dependent on the VLM used in the planner. However, we will not make a recommendation on which VLM to use, since the landscape of VLMs is under continuous development. 

\section{Conclusion and Limitations}
In this work, we design and conduct controlled experiments to investigate how to use VLMs as closed-loop high-level planners. We found that closed-loop high-level planning is preferred over open-loop, even in static environments, due to the limited geometric and logical reasoning capabilities of VLMs. We also found that the control horizon does not substantially impact the performance of VLM high-level planners, given that warm-starting is properly utilized. 

While our work demonstrates promise, the methodology has several limitations that point
to important future research directions. The first is that we use VLMs as \emph{zero-shot} planners, and we recognize that how VLMs are trained (e.g., training data distribution and training procedures) could have profound impacts on the performance of the VLM planner. Furthermore, our experiments do not consider the potential impact that prompt variability could have on the results, though in our initial experiments we see very little difference in the results when we use different prompts, as long as necessary information (ones that are present in the experiment prompts) are provided.

Lastly, the action primitives used in this work are limited to \texttt{pick} and \texttt{place}, but we do expect the observations from this work to apply to VLM high-level planning with other action primitives, since in our experiment setup we make no assumptions about the action primitives, and the VLM high-level planner would treat other action primitives no differently from the \texttt{pick} and \texttt{place} used in our experiments.

\section*{Acknowledgment}
We would like to thank the authors of the work \cite{wang2024_llm3} for their open-source repository, as we developed our experiments based on their repository.

\bibliographystyle{IEEEtran}
\bibliography{refs} 

\clearpage
\onecolumn
\appendices
\section{Planner Prompts}\label{appen:planner_prompts}

The prompt for open-loop planning is as follows:

\begin{tcolorbox}[colback=white, colframe=black, boxrule=0.5pt, breakable, title=Prompt for Open-Loop Planner]
\begin{quote}
\begin{lstlisting}

You are an AI robot planner that generates a sequence of actions to complete the given task. You are provided with:
1. A description of the task environment and the task itself.
2. The current state and observation of the environment.

Your objective is to reason through the task and generate a sequence of actions from scratch that successfully complete the task. 

The descriptions of the task environment and the task are: 
{domain_desc}

Please generate output step by step, which includes:
1. Reasoning: Your reasoning and analysis of how you generate the sequence of actions based on the task and environment descriptions. Make sure to reason about ALL the  logical constraints as well as the geometric constraints (i.e. whether the boxes are fully inside the basket and if the boxes collide with each other). Please also reason what you see from the two images.
2. Full Plan: A full sequence of actions, starting from the current state and ending at task completion.
Please organize the output following the json format below:
{
    "Reasoning": "My reasoning and strategy for generating this sequence of actions is ...",
    "Full Plan": ["pick(['red_box'], {})", "place(['red_box'], {'x': 0.51, 'y': 0.02, 'theta': 0.00})", ...]
}

Your output in json is (please don't output ```json):

\end{lstlisting}
\end{quote}
\end{tcolorbox}

The prompt for closed-loop planning is as follows:

\begin{tcolorbox}[colback=white, colframe=black, boxrule=0.5pt, breakable, title=Prompt for Closed Loop Planner]
\begin{quote}
\begin{lstlisting}

You are an AI robot planner that generates a sequence of actions to complete the given task, using a Model Predictive Control (MPC) style of planning: you are asked to re-generate/re-plan a complete sequence of actions that would complete the task from the current state every time the robot has executed {n_steps_replan} actions of the previously generated sequence of actions, or one of the actions from the previously generated sequence of action fail to execute, and you are provided with:
1. A description of the task environment and the task itself.
2. The current state and observation of the environment.
3. The previously generated sequence of actions along with the execution status of each action.

Your objective is to reason through the task and the previously generated sequence of actions, and generate a complete sequence of actions that would successfully complete the task. You can generate the new sequence of actions by modifying the previously generated sequence of actions. At this moment, you are tasked to generate a complete sequence of actions that would complete the task due to one of the following reasons:
1. A fixed number (={n_steps_replan}) of actions from the previously generated sequence of actions have been executed successfully, but the task is not completed yet. In this case, the remaining actions are not executed by design of the MPC planning style as you are asked to replan. You should reason through the remaining actions in the previously generated sequence and decide if you need to modify the sequence. If the remaining action would complete task while respecting task constraints, you don't need to modify the sequence, and the new sequence is simply the remaining actions in the previously generated sequence. 
2. One of the actions from the previously generated sequence of action fail to execute before the task is completed. In this case, you should reason through why the specific action fails to execute and potentially adjust the remaining actions such that the task can be completed successfully while respecting task constraints. 

The descriptions of the task environment and the task are: 
{domain_desc}

The previously generated sequence of action is:
{prev_plan}

Please generate output step by step, which includes:
1. Reasoning: Your reasoning and analysis of how you generate the sequence of actions based on the task and environment descriptions and the previously generated sequence of actions. Make sure to reason about ALL the logical constraints as well as the geometric constraints (i.e. whether the objects are fully inside the plate/basket and if the objects collide with each other). Please also reason what you see from the two images. 
2. Full Plan: A full sequence of actions, starting from the current state and ending at task completion.
Please organize the output following the json format below:
{
    "Reasoning": "My reasoning and strategy for generating this sequence of actions is ...",
    "Full Plan": ["pick(['apple'], {})", "place(['apple'], {'x': 0.51, 'y': 0.02, 'theta': 0.00})", ...]
}
Your output in json is (please don't output ```json):
    
\end{lstlisting}
\end{quote}
\end{tcolorbox}

The prompt for closed-loop planning with no warm-starting is as follows:

\begin{tcolorbox}[colback=white, colframe=black, boxrule=0.5pt, breakable, title=Prompt for Closed Loop Planner with No Warm-Starting]
\begin{quote}
\begin{lstlisting}

You are an AI robot planner that generates a sequence of actions to complete the given task. You are provided with:
1. A description of the task environment and the task itself.
2. The current state and observation of the environment.

Your objective is to reason through the task and generate a sequence of actions from scratch that successfully complete the task. 

The descriptions of the task environment and the task are: 
{domain_desc}

Please generate output step by step, which includes:
1. Reasoning: Your reasoning and analysis of how you generate the sequence of actions based on the task and environment descriptions. Make sure to reason about ALL the  logical constraints as well as the geometric constraints (i.e. whether the objects are fully inside the plate/basket and if the objects collide with each other). Please also reason what you see from the two images.
2. Full Plan: A full sequence of actions, starting from the current state and ending at task completion.
Please organize the output following the json format below:
{
    "Reasoning": "My reasoning and strategy for generating this sequence of actions is ...",
    "Full Plan": ["pick(['apple'], {})", "place(['apple'], {'x': 0.51, 'y': 0.02, 'theta': 0.00})", ...]
}

Your output in json is (please don't output ```json):

\end{lstlisting}
\end{quote}
\end{tcolorbox}

\newpage
\section{Task Environment Prompts}\label{appen:env_prompts}

The prompt for $\cubeeasy$ is as follows:

\begin{tcolorbox}[colback=white, colframe=black, boxrule=0.5pt,
                  breakable, title=Prompt for $\cubeeasy$]
\begin{quote}
\begin{lstlisting}
The tabletop environment has a robot arm, a basket and several boxes. The robot sits at (-0.65, 0), faces positive x-axis, while positive z-axis points up. The goal is to pack all the boxes into the basket. For every box, make sure its boundary is in the basket.

You must satisfy the following task constraints:
- The blue box must be put into the basket after the red box and brown box is already in the basket.
- The red box must be placed in the basket before the grey box is placed.
- The cyan box must be placed before the red box and black box are placed in the basket.
- The grey box must be placed right before or after the brown box.
- The green box must be placed before the brown box and after the cyan box.
- The black box and cyan box cannot be placed consecutively.

The robot has the following primitive actions, where each primitive action can take a list of objects and parameters as input:
- pick([obj], {}): pick up obj, with no parameters. You should select the proper obj.
- place([obj], {"x": [-0.5, 0.5], "y": [-0.5, 0.5], "theta": [-3.14, 3.14]}): place obj at location (x, y) with planar rotation theta, where x ranges (-0.5, 0.5), y ranges (-0.5, 0.5), and theta ranges (-3.14, 3.14). You should select the proper obj, the x,y placement of the center of mass of obj, and the rotation theta of obj.
\end{lstlisting}
\end{quote}
\end{tcolorbox}

The prompt for $\ycbeasy$ is as follows:

\begin{tcolorbox}[colback=white, colframe=black, boxrule=0.5pt,
                  breakable, title=Prompt for $\ycbeasy$]
\begin{quote}
\begin{lstlisting}

The tabletop environment has a robot arm, a basket and several objects. The robot sits at (-0.65, 0), faces positive x-axis, while positive z-axis points up. The goal is to pack all the objects into the basket. For every object, make sure its boundary is in the basket.

You must satisfy the following task constraints:
- The master chef can must be put into the basket after the tomato soup can and tuna fish can is already in the basket.
- The tomato soup can must be placed in the basket before the baseball is placed.
- The lemon must be placed before the tomato soup can and orange are placed in the basket.
- The baseball must be placed right before or after the tuna fish can.
- The rubiks cube must be placed before the tuna fish can and after the lemon. 
- The orange and lemon cannot be placed consecutively. 

The robot has the following primitive actions, where each primitive action can take a list of objects and parameters as input:
- pick([obj], {}): pick up obj, with no parameters. You should select the proper obj.
- place([obj], {"x": [-0.5, 0.5], "y": [-0.5, 0.5], "theta": [-3.14, 3.14]}): place obj at location (x, y) with planar rotation theta, where x ranges (-0.5, 0.5), y ranges (-0.5, 0.5), and theta ranges (-3.14, 3.14). You should select the proper obj, the x,y placement of the center of mass of obj, and the rotation theta of obj.

\end{lstlisting}
\end{quote}
\end{tcolorbox}

The prompt for $\ycbmedium$ is as follows:

\begin{tcolorbox}[colback=white, colframe=black, boxrule=0.5pt,
                  breakable, title=Prompt for $\ycbmedium$]
\begin{quote}
\begin{lstlisting}

The tabletop environment has a robot arm, a basket, a plate and several objects, which are intially located in the staging area. The robot sits at (-0.65, 0), faces positive x-axis, while positive z-axis points up. The goal is to sort the objects so that the fruits are on the plate and the other objects are in the basket.

You must satisfy the following task constraints:
- The master chef can must be put into the basket after the tomato soup can and tuna fish can is already in the basket.
- The tomato soup can must be placed in the basket before the baseball is placed in the basket.
- The lemon must be placed on the plate before the tomato soup can is in the basket and before the orange is placed on the plate.
- The baseball must be placed in the basket right before or after the tuna fish can in the basket.
- The softball must be placed in the basket right after the baseball is placed in the basket.
- The rubiks cube must be placed in the basket before the tuna fish can is in the basket and after the lemon is on the plate. 
- The orange and lemon cannot be placed consecutively. 
 
The robot has the following primitive actions, where each primitive action can take a list of objects and parameters as input:
- pick([obj], {}): pick up obj, with no parameters. You should select the proper obj.
- place([obj], {"x": [-0.5, 0.5], "y": [-0.5, 0.5], "theta": [-3.14, 3.14]}): place obj at location (x, y) with planar rotation theta, where x ranges (-0.5, 0.5), y ranges (-0.5, 0.5), and theta ranges (-3.14, 3.14). You should select the proper obj, the x,y placement of the center of mass of obj, and the rotation theta of obj.

\end{lstlisting}
\end{quote}
\end{tcolorbox}

The prompt for $\ycbhard$ is as follows:

\begin{tcolorbox}[colback=white, colframe=black, boxrule=0.5pt,
                  breakable, title=Prompt for $\ycbhard$]
\begin{quote}
\begin{lstlisting}

The tabletop environment has a robot arm, a basket, a plate and several objects, which are initially located in the staging area. The robot sits at (-0.65, 0), faces positive x-axis, while positive z-axis points up. The goal is to make a fruit salad (i.e., putting all the necessary ingredients into the plate). The unused ingredients should be put into the brown basket.

You must satisfy the following task constraints:
- I don't like sour fruits. 
- You should put away the ingredients you are not going to used before making the fruit salad.
- Lemon should not be placed first.
- Strawberry should not be placed last. 
- Pear should be placed before strawberry. 
- Once you placed any object, you should not attempt to place it again.

The robot has the following primitive actions, where each primitive action can take a list of objects and parameters as input:
- pick([obj], {}): pick up obj, with no parameters. You should select the proper obj.
- place([obj], {"x": [-0.5, 0.5], "y": [-0.5, 0.5], "theta": [-3.14, 3.14]}): place obj at location (x, y) with planar rotation theta, where x ranges (-1, 1), y ranges (-1, 1), and theta ranges (-3.14, 3.14). You should select the proper obj, the x,y placement of the center of mass of obj, and the rotation theta of obj.

\end{lstlisting}
\end{quote}
\end{tcolorbox}

\newpage
\section{Comprehensive Experiment Results}\label{appen:detailed_results}

In each of the following tables, we show the Goal Achieved Rate (GAR), Task Completion Rate (TCR), Correct Final Logical Plan Rate (CFP), Positive Logical Correction Rate (PCR), and Negative Logical Correction Rate (NCR). Finally, we show the number of valid trials, where the low-level motion planner performs properly. We present these results for the following planners: 1) closed-loop-full (CL-F), 2) closed-loop-half (CL-H), 3) closed-loop-half-non-warm-starting (CL-H-NWS), 4) closed-loop-short (CL-S), 5) closed-loop-short-non-warm-starting (CL-S-NWS), and 6) open-loop (OL). 

\begin{table}[htbp]
\centering
\caption{Comparison of planner metrics for $\cubeeasy$.}
\label{tab:cubeeasyfull}
\begin{tabular}{ll|cccccc}
\toprule
Model Name & Planner Type & GAR & TCR & CFP & PCR & NCR & \# of Valid Trials \\
\midrule
GPT-4.1-mini & CL-F & 0.720 & 0.620 & 0.840 & 0.000 & 0.000 & 50 \\
 & CL-H & 0.880 & 0.780 & 0.840 & 0.000 & 0.000 & 50 \\
 & CL-H-NWS & 0.583 & 0.479 & 0.780 & 0.000 & 0.027 & 48 \\
 & CL-S & 0.860 & 0.820 & 0.960 & 0.235 & 0.004 & 50 \\
 & CL-S-NWS & 0.795 & 0.773 & 0.980 & 0.765 & 0.011 & 44 \\
 & OL & 0.620 & 0.520 & 0.840 & – & – & 50 \\
\hline
Gemini-2.5-flash & CL-F & 0.380 & 0.340 & 0.900 & 0.200 & 0.000 & 50 \\
 & CL-H & 0.640 & 0.540 & 0.880 & 0.000 & 0.000 & 50 \\
 & CL-H-NWS & 0.500 & 0.420 & 0.820 & 0.091 & 0.038 & 50 \\
 & CL-S & 0.840 & 0.780 & 0.920 & 0.184 & 0.010 & 50 \\
 & CL-S-NWS & 0.180 & 0.180 & 0.900 & 0.500 & 0.049 & 50 \\
 & OL & 0.160 & 0.140 & 0.880 & – & – & 50 \\
\hline
Llama-4-Maverick-17B & CL-F & 0.959 & 0.796 & 0.840 & 0.000 & 0.000 & 49 \\
 & CL-H & 0.800 & 0.660 & 0.820 & 0.000 & 0.014 & 50 \\
 & CL-H-NWS & 0.320 & 0.260 & 0.620 & 0.000 & 0.087 & 50 \\
 & CL-S & 0.327 & 0.306 & 0.900 & 0.053 & 0.001 & 49 \\
 & CL-S-NWS & 0.000 & 0.000 & 0.760 & 0.081 & 0.021 & 50 \\
 & OL & 0.857 & 0.714 & 0.840 & – & – & 49 \\
\bottomrule
\end{tabular}
\end{table}

\begin{table}[htbp]
\centering
\caption{Comparison of planner metrics for $\ycbeasy$.}
\label{tab:ycbeasyfull}
\begin{tabular}{ll|cccccc}
\toprule
Model Name & Planner Type & GAR & TCR & CFP & PCR & NCR & \# of Valid Trials \\
\midrule
GPT-4.1-mini & CL-F & 0.646 & 0.646 & 1.000 & – & 0.000 & 48 \\
 & CL-H & 0.776 & 0.694 & 0.820 & 0.000 & 0.093 & 49 \\
 & CL-H-NWS & 0.750 & 0.500 & 0.660 & 0.000 & 0.157 & 44 \\
 & CL-S & 0.957 & 0.894 & 0.920 & 0.000 & 0.008 & 47 \\
 & CL-S-NWS & 1.000 & 0.812 & 0.820 & 0.000 & 0.019 & 48 \\
 & OL & 0.400 & 0.400 & 1.000 & – & – & 50 \\
\hline
Gemini-2.5-flash & CL-F & 0.760 & 0.720 & 0.960 & – & 0.000 & 50 \\
 & CL-H & 0.980 & 0.918 & 0.900 & 0.000 & 0.039 & 49 \\
 & CL-H-NWS & 0.800 & 0.240 & 0.340 & 0.000 & 0.277 & 50 \\
 & CL-S & 0.959 & 0.939 & 0.980 & 0.200 & 0.002 & 49 \\
 & CL-S-NWS & 0.833 & 0.479 & 0.600 & 0.045 & 0.043 & 48 \\
 & OL & 0.580 & 0.540 & 0.960 & – & – & 50 \\
\hline
Llama-4-Maverick-17B & CL-F & 0.200 & 0.180 & 0.840 & 0.000 & 0.000 & 50 \\
 & CL-H & 0.283 & 0.239 & 0.760 & 0.000 & 0.030 & 46 \\
 & CL-H-NWS & 0.255 & 0.128 & 0.480 & 0.000 & 0.133 & 47 \\
 & CL-S & 0.435 & 0.370 & 0.800 & 0.057 & 0.012 & 46 \\
 & CL-S-NWS & 0.271 & 0.125 & 0.660 & 0.121 & 0.038 & 48 \\
 & OL & 0.120 & 0.100 & 0.840 & – & – & 50 \\
\bottomrule
\end{tabular}
\end{table}

\begin{table}[htbp]
\centering
\caption{Comparison of planner metrics for $\ycbmedium$.}
\label{tab:ycbmediumfull}
\begin{tabular}{ll|cccccc}
\toprule
Model Name & Planner Type & GAR & TCR & CFP & PCR & NCR & \# of Valid Trials \\
\midrule
GPT-4.1-mini & CL-F & 0.956 & 0.089 & 0.100 & 0.000 & 0.000 & 45 \\
 & CL-H & 0.909 & 0.114 & 0.100 & 0.000 & 0.029 & 44 \\
 & CL-H-NWS & 0.921 & 0.132 & 0.120 & 0.173 & 0.309 & 38 \\
 & CL-S & 0.923 & 0.179 & 0.180 & 0.017 & 0.002 & 39 \\
 & CL-S-NWS & 0.925 & 0.325 & 0.260 & 0.161 & 0.107 & 40 \\
 & OL & 0.688 & 0.083 & 0.100 & – & – & 48 \\
\hline
Gemini-2.5-flash & CL-F & 0.708 & 0.646 & 0.920 & 0.000 & 0.000 & 48 \\
 & CL-H & 0.854 & 0.771 & 0.920 & 0.000 & 0.021 & 48 \\
 & CL-H-NWS & 0.359 & 0.359 & 0.780 & 0.000 & 0.082 & 39 \\
 & CL-S & 0.958 & 0.896 & 0.920 & 0.167 & 0.012 & 48 \\
 & CL-S-NWS & 0.667 & 0.500 & 0.780 & 0.214 & 0.044 & 24 \\
 & OL & 0.300 & 0.240 & 0.920 & – & – & 50 \\
\hline
Llama-4-Maverick-17B & CL-F & 0.604 & 0.458 & 0.720 & 0.000 & 0.000 & 48 \\
 & CL-H & 0.717 & 0.500 & 0.700 & 0.000 & 0.009 & 46 \\
 & CL-H-NWS & 0.020 & 0.020 & 0.640 & 0.229 & 0.095 & 49 \\
 & CL-S & 0.864 & 0.591 & 0.740 & 0.066 & 0.012 & 44 \\
 & CL-S-NWS & 0.000 & 0.000 & 0.360 & 0.123 & 0.095 & 42 \\
 & OL & 0.080 & 0.020 & 0.720 & – & – & 50 \\
\bottomrule
\end{tabular}
\end{table}

\begin{table}[htbp]
\centering
\caption{Comparison of planner metrics for $\ycbhard$.}
\label{tab:ycbhardfull}
\begin{tabular}{ll|cccccc}
\toprule
Model Name & Planner Type & GAR & TCR & CFP & PCR & NCR & \# of Valid Trials \\
\midrule
GPT-4.1-mini & CL-F & 0.814 & 0.093 & 0.140 & 0.118 & 0.130 & 43 \\
 & CL-H & 0.783 & 0.109 & 0.180 & 0.167 & 0.087 & 46 \\
 & CL-H-NWS & 0.250 & 0.000 & 0.220 & 0.208 & 0.053 & 40 \\
 & CL-S & 0.541 & 0.135 & 0.220 & 0.118 & 0.075 & 37 \\
 & CL-S-NWS & 0.020 & 0.000 & 0.120 & 0.128 & 0.088 & 49 \\
 & OL & 0.540 & 0.040 & 0.160 & – & – & 50 \\
\hline
Gemini-2.5-flash & CL-F & 0.880 & 0.260 & 0.280 & 0.091 & 0.200 & 50 \\
 & CL-H & 1.000 & 0.640 & 0.640 & 0.417 & 0.055 & 50 \\
 & CL-H-NWS & 0.760 & 0.540 & 0.600 & 0.442 & 0.125 & 50 \\
 & CL-S & 1.000 & 0.740 & 0.740 & 0.346 & 0.111 & 50 \\
 & CL-S-NWS & 0.860 & 0.600 & 0.600 & 0.438 & 0.169 & 50 \\
 & OL & 0.700 & 0.240 & 0.320 & – & – & 50 \\
\hline
Llama-4-Maverick-17B & CL-F & 0.020 & 0.000 & 0.340 & 0.296 & 0.265 & 49 \\
 & CL-H & 0.020 & 0.020 & 0.320 & 0.298 & 0.228 & 49 \\
 & CL-H-NWS & 0.040 & 0.000 & 0.440 & 0.354 & 0.180 & 50 \\
 & CL-S & 0.070 & 0.000 & 0.220 & 0.231 & 0.180 & 43 \\
 & CL-S-NWS & 0.000 & 0.000 & 0.500 & 0.390 & 0.151 & 50 \\
 & OL & 0.000 & 0.000 & 0.440 & – & – & 49 \\
\bottomrule
\end{tabular}
\end{table}

\end{document}